%% file: colm2026_conference.tex
\PassOptionsToPackage{table,xcdraw}{xcolor}
\documentclass{article} 
\usepackage[preprint]{colm2026_conference}

\usepackage{microtype}
\usepackage{inconsolata}
\usepackage{soul}   
\usepackage{hyperref}
\usepackage{url}
\usepackage{booktabs}
\usepackage{graphicx}
\usepackage{amsmath}
\usepackage{amssymb}
\usepackage{multirow}
\usepackage[table,xcdraw]{xcolor}
\usepackage[normalem]{ulem}
\useunder{\uline}{\ul}{}
\usepackage{booktabs}
\usepackage[ruled,lined]{algorithm2e}

\usepackage[most]{tcolorbox}
\tcbuselibrary{listings,skins,breakable}
\usepackage{listings} 

\usepackage{booktabs,makecell,xcolor,pifont,colortbl}
\usepackage{latex/daiwei_style}

\usepackage{lineno}

\definecolor{darkblue}{rgb}{0, 0, 0.5}
\hypersetup{colorlinks=true, citecolor=darkblue, linkcolor=darkblue, urlcolor=darkblue}

\title{Grounded Token Initialization for New Vocabulary in LMs for Generative Recommendation}

\author{Daiwei Chen$^{1,2}$\thanks{Work done during internship at LinkedIn Corporation. Correspondence to: \texttt{dchen365@wisc.edu}}\quad
Zhoutong Fu$^{2}$,\quad
Chengming Jiang$^{2}$,\quad
Haichao Zhang$^{3}$,\quad
Ran Zhou$^{2}$
\\
\bf
Tan Wang$^{2}$,\quad
Chunnan Yao$^{2}$,\quad
Guoyao Li$^{2}$,\quad
Rui Cai$^{4}$,\quad
Yihan Cao$^{2}$,\quad
Ruijie Jiang$^{2}$,
\\
\bf
Fedor Borisyuk$^{2}$,\quad
Jianqiang Shen$^{2}$,\quad
Jingwei Wu$^{2}$,\quad
Ramya Korlakai Vinayak$^{1}$
\\[6pt]
\normalfont\small
\makebox[\dimexpr\textwidth-2\tabcolsep\relax][c]{%
$^{1}$University of Wisconsin-Madison\quad
$^{2}$LinkedIn Corporation}
\\
\makebox[\dimexpr\textwidth-2\tabcolsep\relax][c]{%
$^{3}$Northeastern University\quad
$^{4}$University of California, Davis}
}

\DeclareRobustCommand{\fancyname}{\texttt{\textbf{GTI}}}

\begin{document}

\ifcolmsubmission
\linenumbers
\fi

\maketitle
\vspace{-15.121pt} 

\input{sections/00abstract}

\input{sections/01intro}

\input{sections/02issue}

\input{sections/03method}

\input{sections/04results}

\input{sections/05related_work}

\input{sections/06conclusion}

\input{sections/07acknowledgment}

\bibliography{colm2026_conference}
\bibliographystyle{colm2026_conference}

\input{sections/99appendix}

\end{document}

%% file: sections/00abstract.tex
\begin{abstract}
Language models (LMs) are increasingly extended with new learnable vocabulary tokens for domain-specific tasks, such as Semantic-ID tokens in generative recommendation. The standard practice initializes these new tokens as the mean of existing vocabulary embeddings, then relies on supervised fine-tuning to learn their representations. We present a systematic analysis of this strategy: through spectral and geometric diagnostics, we show that mean initialization collapses all new tokens into a degenerate subspace, erasing inter-token distinctions that subsequent fine-tuning struggles to fully recover. These findings suggest that \emph{token initialization} is a key bottleneck when extending LMs with new vocabularies.
Motivated by this diagnosis, we propose the \emph{Grounded Token Initialization Hypothesis}: linguistically grounding novel tokens in the pretrained embedding space before fine-tuning better enables the model to leverage its general-purpose knowledge for novel-token domains. We operationalize this hypothesis as \fancyname\ (Grounded Token Initialization), a lightweight grounding stage that, prior to fine-tuning, maps new tokens to distinct, semantically meaningful locations in the pretrained embedding space using only paired linguistic supervision. Despite its simplicity, \fancyname\ outperforms both mean initialization and existing auxiliary-task adaptation methods in the majority of evaluation settings across multiple generative recommendation benchmarks, including industry-scale and public datasets. Further analyses show that grounded embeddings produce richer inter-token structure that persists through fine-tuning, corroborating the hypothesis that initialization quality is a key bottleneck in vocabulary extension.
\end{abstract}

%% file: sections/01intro.tex
\section{Introduction}
Pretrained language models (LMs) are increasingly adapted to specialized domains by extending their vocabulary with new learnable tokens. A prominent example is generative retrieval, where items~\citep{TIGER, GRreview} or documents~\citep{tay2022transformer} are assigned discrete semantic codes and generated autoregressively by the LM; similar challenges arise whenever domain-specific symbols must be integrated into a pretrained vocabulary. These systems introduce thousands of new tokens into the model's vocabulary, and a fundamental challenge is how to incorporate them into the pretrained embedding space so that the LM can transfer its general-purpose knowledge to the novel-token domain.

The prevailing practice initializes new token embeddings as the mean of the existing vocabulary embeddings~\citep{hewitt2021initializing}. This heuristic is widely adopted because it is simple, places new tokens on the pretrained embedding manifold, and provides a tighter KL-divergence upper bound on output probabilities. However, it collapses all new tokens into a single point in embedding space, erasing inter-token distinctions and stripping domain-level semantics. An existing alternative~\citep{LC-Rec} employs auxiliary-task adaptation of the full LM to induce linguistic signals for new tokens, but the multi-task training introduces an objective mismatch: the auxiliary losses are not aligned with the target downstream task, resulting in limited and inconsistent gains.

In this paper, we identify token-embedding misalignment as a fundamental limitation when extending pretrained LMs with new vocabulary, and propose the \emph{Grounded Token Initialization Hypothesis}: \textbf{linguistically grounding novel tokens in an LM's pretrained embedding space, before fine-tuning, better enables the model to leverage its general-purpose knowledge for novel-token domains.}
The intuition is that pretrained LM embeddings encode rich linguistic structure, semantically related tokens occupy nearby regions~\citep{levy2014neural}, and the model's attention and feed-forward layers have learned to exploit this geometry~\citep{gao2019representation}. If new tokens are placed meaningfully within this structure, the LM can immediately leverage its existing representations to process them in context, rather than relying on fine-tuning alone to recover from a degenerate starting point. This motivates framing vocabulary extension as a token-grounding problem: new token embeddings should be grounded in linguistically meaningful representations while remaining coherent with the pretrained LM's embedding geometry.

\begin{figure}[]
    \centering
    \vspace{-30pt}
    \includegraphics[width=1.\linewidth]{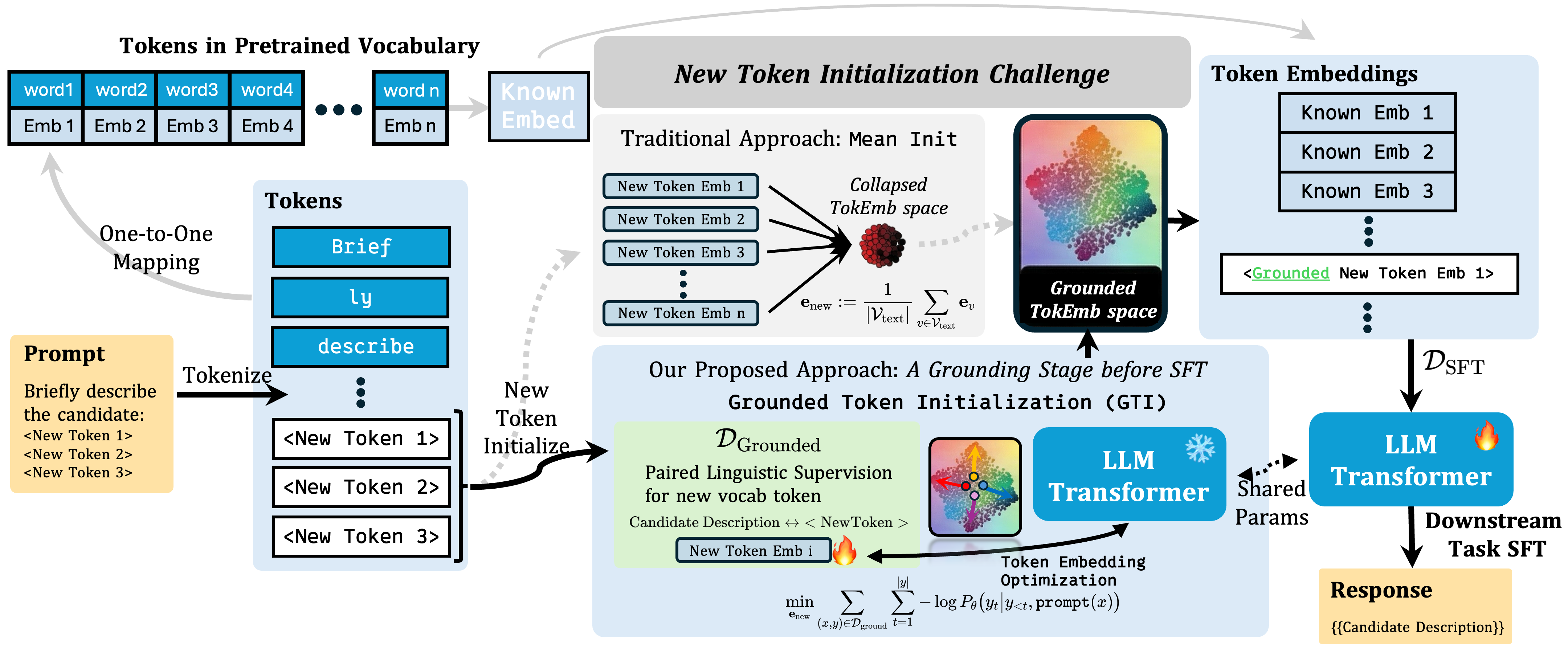}
    \vspace{-15pt}
    \caption{Overview of the \fancyname\ grounding stage. The LM backbone and original vocabulary embeddings are frozen (snowflake); only the newly introduced Semantic-ID (SID) token embeddings (\(|\mathcal{V}_{\mathrm{SID}}|\!\times\! D\) parameters, fire) are trained. Paired prompts map between natural language descriptions and SID tokens in both directions, grounding the new tokens in the pretrained embedding space. This stage is inserted before standard end-to-end fine-tuning (see Section~\ref{sec:method}).}
    \vspace{-20pt}
    \label{fig:framework}
\end{figure}

Building on this hypothesis, we introduce \fancyname, a simple and effective grounded token initialization method. Before downstream fine-tuning, \fancyname\ freezes the LM backbone and grounds newly introduced token embeddings using paired supervision between natural language descriptions and the corresponding new tokens  (Figure~\ref{fig:framework}). This grounding stage resolves the mismatch between well-trained vocabulary embeddings and newly initialized tokens, providing the LM with a semantically structured starting point for subsequent end-to-end fine-tuning of the full model for target downstream tasks.

We validate \fancyname\ on Generative Recommendation (GR)~\citep{TIGER, pmlr-v235-zhai24a}, a challenging and practically important application of vocabulary extension. GRs have attracted growing attention in both academia and industry~\citep{ding2026doesgenerativerecommendationgeneralize, mtgr, oneSearch, OneRec}, as they dramatically simplify retrieval by autoregressively generating item identifiers token-by-token from user interaction histories, replacing the expensive user--item inner products required by traditional dense-embedding methods~\citep{MF, NCF, LightGCN, NGCF}. GRs can further exploit scaling-law behavior as model size and data increase~\citep{mtgr}, offering a clear path to continued improvement. The GR setting is a particularly demanding testbed for grounded token initialization: large sets of new learnable Semantic-ID (SID) tokens must be incorporated into pretrained LMs, each encoding fine-grained item-level semantics and hierarchical codebook structure that should be properly grounded in the LM's embedding space to support effective retrieval.

\newpage
\textbf{Contributions.}
\vspace{-7pt}
\begin{enumerate}
    \item \textbf{Diagnosis.} Through spectral and geometric analysis, we characterize the token-embedding misalignment caused by mean initialization: all new learnable tokens collapse into a degenerate, low-rank subspace that does not fully recover under subsequent fine-tuning. This motivates the \emph{Grounded Token Initialization Hypothesis}: linguistically grounding new tokens before fine-tuning better enables the LM to leverage its pretrained knowledge for the new domain.
    \item \textbf{Methodology.} We introduce \fancyname, a simple and effective grounding stage that freezes the LM backbone and learns new token embeddings via paired linguistic supervision before standard fine-tuning, providing a semantically structured starting point for downstream adaptation.
    \item \textbf{Finding.} On generative recommendation benchmarks, spanning industry-scale and public datasets, \fancyname\ consistently outperforms both direct supervised fine-tuning and \texttt{LC-Rec}~\citep{LC-Rec}, an existing approach that jointly adapts the full model via auxiliary tasks. These results suggest that token initialization is a key bottleneck in vocabulary extension.
\end{enumerate}

%% file: sections/02issue.tex
\section{Token-Embedding Misalignment}
\label{subsect:misalign}

We formalize the vocabulary extension problem in the context of generative retrieval, our primary application domain, and then use spectral and geometric diagnostics to characterize a systematic token-embedding misalignment that arises from standard initialization practices when new tokens are added to a pretrained language model.

\paragraph{Generative Retrieval.}
We adopt the framework of \citet{TIGER}. Each item $I_i \in \mathcal{I}$ has content features (title, description, etc.) that a pretrained text encoder maps to a semantic embedding $\mathbf{z}_i \in \mathbb{R}^d$. An RQ-VAE~\citep{RQ-VAE} with $L$ codebook levels of $K$ entries each discretizes $\mathbf{z}_i$ into a Semantic ID $(c_1,\ldots,c_L)$, $c_l \in \{1,\ldots,K\}$, via recursive residual quantization:
\[
\mathbf{r}_{1} := \mathbf{z}_i;\qquad c_l = \arg\min_{k}\bigl\|\mathbf{r}_l - \mathbf{q}_k^{(l)}\bigr\|_2, \quad \mathbf{r}_{l+1} := \mathbf{r}_l - \mathbf{q}_{c_l}^{(l)}, \quad l = 1,\ldots,L,
\]
where $\bigl\{\mathbf{q}_k^{(l)}\bigr\}_{k=1}^{K}\!\subset\mathbb{R}^d$ is the level-$l$ codebook. The $K \!\times\! L$ SID codes\footnote{SID tokens are labeled by level to encode codebook membership, e.g., \texttt{<a\_1>, <b\_1>, <c\_1>.}.} are appended to the LM's original vocabulary $\mathcal{V}_{\mathrm{text}}$ as new tokens $\mathcal{V}_{\mathrm{SID}}$. Given a context $\mathbf{x}$, either a user's interaction history (retrieval) or a natural language query (search), the LM generates the target Semantic ID autoregressively:
\[
P_\theta(c_1,\ldots,c_L \mid \mathbf{x}) = \prod_{t=1}^{L} P_\theta(c_t \mid c_{<t}, \mathbf{x}).
\]

\paragraph{Mean-of-Vocabulary Initialization.} Standard practice initializes all novel token embeddings to the mean of the existing vocabulary embeddings~\citep{hewitt2021initializing}:
\begin{equation}
\label{eq:mean-init}
\mathbf{e}_{c} := \frac{1}{|\mathcal{V}_{\mathrm{text}}|}\sum_{v \in \mathcal{V}_{\mathrm{text}}} \mathbf{e}_v, \quad \forall\; c \in \mathcal{V}_{\mathrm{SID}},
\end{equation}
where $\mathbf{e}_v$ denotes the input embedding of token $v$.

\begin{figure}[h!]
    \centering
    \vspace{-30pt}
    \makebox[\linewidth][c]{%
        \includegraphics[width=1.1\linewidth]{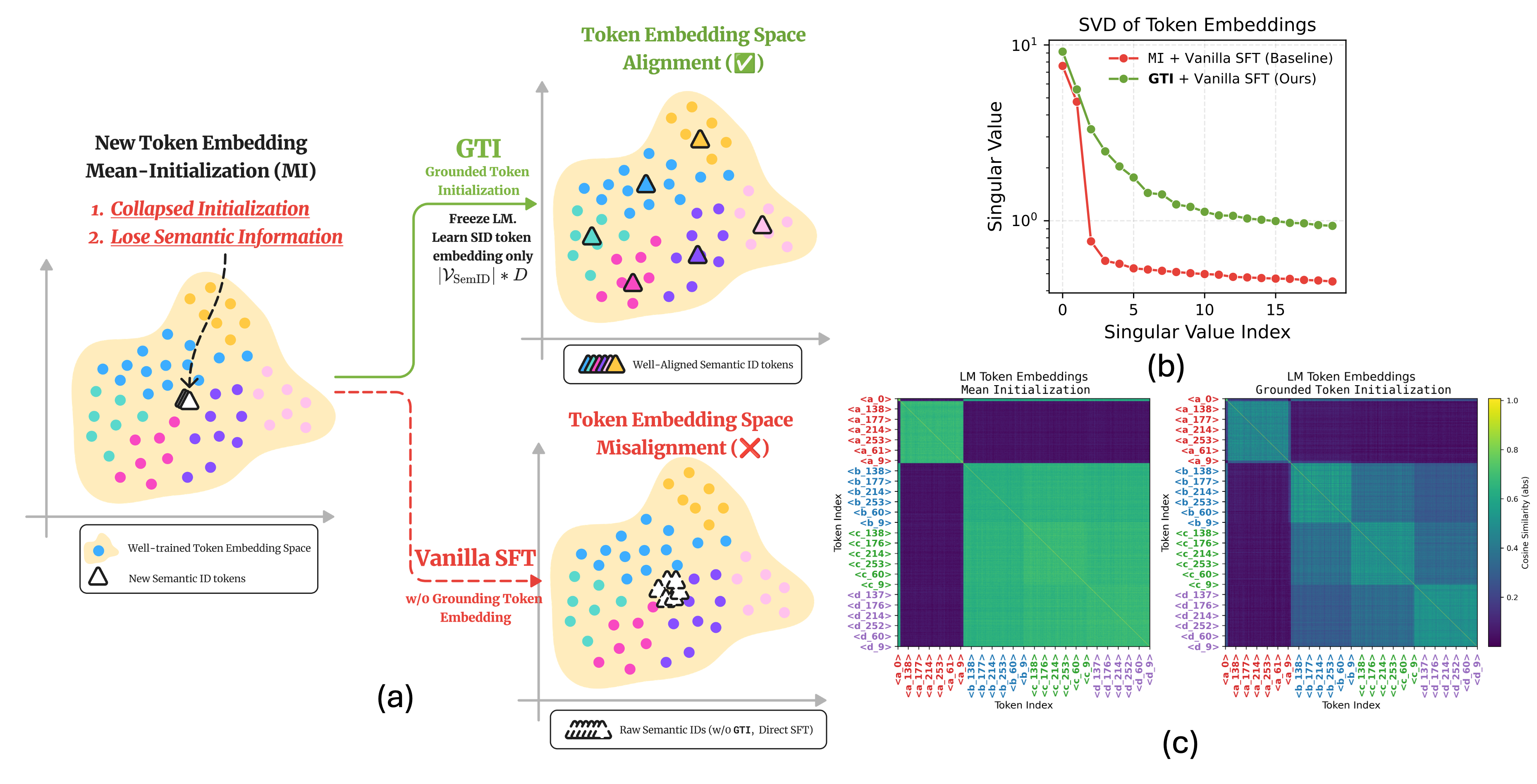}
    }
    \vspace{-20pt}
    \caption{\textbf{Token-embedding collapse under mean initialization and the effect of grounding.} (a)~\emph{Left:} Mean initialization maps all SID tokens (white triangles) to a single point, collapsing inter-token distinctions. \emph{Top-right:} \fancyname\ grounds SID tokens (colored triangles) into distinct regions by training only the $|\mathcal{V}_{\mathrm{SID}}|\!\times\!d$ embedding parameters while freezing the backbone. \emph{Bottom-right:} Fine-tuning without grounding does not fully resolve the collapse (see Figure~\ref{fig:svd_rsa}). (b)\&(c)~\fancyname\ initialization yields higher effective rank and preserves blockwise hierarchical structure among SID tokens after downstream task supervised finetuning.}
    \vspace{-16pt}
    \label{fig:misalignment_diagram}
\end{figure}

\paragraph{Diagnosing the misalignment.}
Under mean-of-vocabulary initialization (Eq.~\ref{eq:mean-init}), every new token receives an identical embedding, 1) collapsing all inter-token distinctions and 2) discarding the semantic structure each token should encode (Fig.~\ref{fig:misalignment_diagram}, left). This heuristic is nonetheless widely adopted~\citep{wolf2020transformers} because it places new tokens on the pretrained manifold and yields a tighter KL-divergence upper bound on output probabilities compared with random initialization~\citep{hewitt2021initializing}. Random initialization, conversely, assigns distinct vectors to each token but places them without coherent relation to the pretrained manifold, providing no linguistic prior for the model to build on. Pairwise cosine similarities among token embeddings (Fig.~\ref{fig:init_cossim}) confirm that mean initialization produces a near-uniform similarity block across all SID tokens, while random initialization yields unstructured noise.

We examine whether supervised fine-tuning recovers the structure lost under mean initialization. The pairwise similarity among SID embeddings (Fig.\ref{fig:misalignment_diagram} (c) and Fig.\ref{fig:pairwise_sft} Left\&Mid) and the singular-value decomposition of the SID embedding matrix $E_{\mathrm{SID}} \in \mathbb{R}^{|\mathcal{V}_{\mathrm{SID}}|\times d}$ after supervised fine-tuning from the mean-initialized state (Fig.\ref{fig:misalignment_diagram} (b) and Fig.\ref{fig:svd_rsa}) reveals rapid spectral decay and low effective rank, confirming that supervised fine-tuning alone does not recover the inter-token structure lost at mean or random initialization. Taken together, these analyses show that neither strategy provides a suitable starting point: mean initialization places tokens on the pretrained manifold but erases discrimination, while random initialization preserves discrimination but lacks linguistic grounding.
 
\paragraph{Grounded Token Initialization (\texttt{GTI}) Hypothesis.}
These observations motivate our central hypothesis: \textbf{linguistically grounding novel tokens in an LM's pretrained embedding space, before downstream fine-tuning, better enables the model to leverage its general-purpose knowledge for novel-token domains.} Rather than relying on fine-tuning alone to recover from a degenerate initialization, we propose inserting a simple and efficient grounding stage that learns new token embeddings via linguistic supervision with the backbone frozen, before proceeding to standard end-to-end fine-tuning. We operationalize this hypothesis in Section~\ref{sec:method} and verify its effectiveness empirically in Section~\ref{sec:results}.

%% file: sections/03method.tex
\section{\fancyname: Grounded Token Initialization Stage}
\label{sec:method}
The diagnosis in Section~\ref{subsect:misalign} motivates a straightforward modification to the standard training pipeline: before downstream fine-tuning, insert a grounding stage that freezes the LM backbone and only learns new token embeddings via paired linguistic supervision. This design builds on the established principle of training new token embeddings within a frozen LM~\citep{ToolkenGPT, YoLLaVA}. We term the resulting procedure \fancyname. Despite its simplicity, we show that this additional stage yields consistent improvements across multiple benchmarks, including both public and industry-scale datasets (Section~\ref{sec:results}), suggesting that token initialization is a key bottleneck in vocabulary extension.

\paragraph{Algorithm.}
Let $\mathcal{V}=\mathcal{V}_{\text{text}}\cup\mathcal{V}_{\text{new}}$ denote the extended vocabulary, where $\mathcal{V}_{\text{new}}$ are the newly added domain tokens. Given a pretrained autoregressive LM with input-embedding matrix $E\in\mathbb{R}^{|\mathcal{V}|\times d}$, we partition $E$ into the pretrained rows $E_{\text{text}}$ and the new rows $E_{\text{new}}\in\mathbb{R}^{|\mathcal{V}_{\text{new}}|\times d}$. Each domain entity is associated with a natural-language description $x_i$ (e.g., title or definition) and a canonical new-token sequence $y_i=(c_{i,1},\dots,c_{i,L})$. We instantiate \fancyname\ in the generative recommendation setting, where $\mathcal{V}_{\text{new}}=\mathcal{V}_{\text{SID}}$, $x_i$ is an item title/description, and $y_i$ is the corresponding SID sequence.

We construct a grounding corpus $\mathcal{D}_{\mathrm{ground}}=\{(x_i,y_i)\}_{i=1}^n$ pairing each description with its new token sequence, along with reversed pairs $\{(y_i,x_i)\}$ that require the model to generate descriptions from new tokens\footnote{Bidirectional training encourages new token embeddings to encode semantics in both the input and output directions; see ablation in Section~\ref{sec:analysis} and template details in Appendix~\ref{app:prompt-template}.}. Using an instruction-style prompt template $\texttt{prompt}(x)$ (Listing as follows), we minimize the negative log-likelihood over $E_{\text{new}}$:
\begin{equation}
\label{eq:gen}
\min_{E_{\text{new}}}\sum_{(x,y)\in\mathcal{D}_{\text{ground}}}\sum_{t=1}^{|y|}
-\log P_\theta\big(y_t \big| y_{<t},\texttt{prompt}(x)\big)
\end{equation}
where $\theta$ denotes all LM parameters. During this stage, all parameters except $E_{\text{new}}$ are held fixed, including $E_{\text{text}}$ and the LM head, which shares weights with $E$ via the standard tied-embedding parameterization. This weight tying means the grounding stage simultaneously shapes how the model reads and generates new tokens. After grounding, we retain the learned $E_{\text{new}}$ as initialization and proceed with standard supervised fine-tuning of all model parameters $\theta$. Implementation details are provided in Algorithm~\ref{alg:gti-grounding}.

\begin{promptboxgray}[label={lst:semid-prompt-in-paper}]{Item Title/Description $\rightarrow$ Semantic IDs \textit{(Text$\rightarrow$New Vocabulary Tokens)}}
<system>
You are a helpful assistant.
<user>
What item is called {{title}} and described as {{description}}?
<assistant>
{{ITEM SEMANTIC_ID}}
\end{promptboxgray}

%% file: sections/04results.tex
\section{Experiments}
\label{sec:results}
We evaluate \fancyname\ within the highly demanding domain of generative recommendation. This domain serves as an ideal testbed for the initialization bottleneck, as it requires incorporating thousands of new Semantic-ID (SID) tokens into a pretrained language model. To empirically validate whether aligning these tokens with the model’s pre-existing linguistic geometry can prevent semantic collapse, we evaluate across two diverse environments: an industrial-scale candidate retrieval system and the public Vibrent Clothes Rental benchmark.

\subsection{Setup}
\label{sec:setup}

\paragraph{Datasets.} We evaluate across two distinct scales and domains.

\textbf{(1) Industrial candidate retrieval.}\footnote{Data access and use complied with internal privacy and security frameworks. Member profile attributes were processed in accordance with applicable member controls and visibility settings, and analyses were conducted on de‑identified datasets with results reported in aggregate only.} This dataset consists of job requirement--candidate pairs from a world-leading recruitment platform. Each pair is categorized into three relevance levels (good, good\&maybe, and not match) by an internal LLM evaluator according to how many job requirements a candidate satisfies. Due to data-sharing constraints, we report only relative performance gains over the SFT baseline for this dataset.

\textbf{(2) Vibrent Clothes Rental.} To validate generalizability, we adapt the public Vibrent Clothes Rental Dataset~\citep{vibrent_clothes_rental_dataset} into a generative retrieval task, treating users as queries and clothing items as candidates based on historical rental transactions.

\paragraph{Baselines.} 
To strictly isolate the initialization bottleneck, all methods share an identical Qwen3-0.6B backbone and RQ-VAE tokenization structure, differing only in how they introduce novel tokens. 

(1)~\texttt{Vanilla \textbf{SFT}}: New SID tokens are mean-initialized (Eq.~\ref{eq:mean-init}), inducing a semantic collapse. The model relies entirely on downstream fine-tuning to disambiguate tokens from this degenerate starting point.

(2)~\texttt{\textbf{LC-Rec}}~\citep{LC-Rec}: A recent multi-task approach that begins from the same collapsed state but attempts to recover semantic structure by applying auxiliary natural language alignment objectives \textit{during} fine-tuning.

(3)~\texttt{\textbf{GTI\ (Ours)}}: Using the grounding stage described in Section~\ref{sec:method}, we ground the new SID tokens into distinct, semantically meaningful regions of the frozen LM's embedding space, providing a structurally rich starting point for the subsequent SFT procedure.

\paragraph{Evaluation Metrics.} We measure retrieval accuracy using Top-$K$ Precision, Recall, and NDCG. For the industrial dataset, we sample 200 jobs as evaluation queries (retrieving 200 candidates each). To comply with data-sharing constraints, we isolate the direct performance uplift of our grounding stage by reporting results strictly as a relative percentage gain over the standard SFT baseline, formulated as $(M_{\text{method}} - M_{\text{Baseline}}) / M_{\text{Baseline}}$. For the public Vibrent dataset, we adopt the standard leave-one-out sequence splitting strategy~\citep{SelfAttnSeqRec,P5}.

\useunder{\uline}{\ul}{}
\begin{table*}[b!]
\vspace{-20pt}
\caption{Relative Precision@K gain (\%) over SFT baseline on a real-world candidate retrieval dataset. Bold and underline denote the best result.}
\vspace{5pt}
\centering
\makebox[\linewidth][c]{%
\resizebox{1.1\linewidth}{!}{
\begin{tabular}{cl|ccccc|ccccc}
\toprule
\multicolumn{2}{c|}{\multirow{2}{*}{Methodology}} & \multicolumn{5}{c|}{Precision@K (Good Match)} & \multicolumn{5}{c}{Precision@K (Good \& Maybe Match)} \\
\multicolumn{2}{c|}{} & P@5 & P@10 & P@20 & P@50 & P@100 & P@5 & P@10 & P@20 & P@50 & P@100 \\ 
\midrule

\multicolumn{2}{c|}{\texttt{MI}+Vanilla SFT (\texttt{Baseline})} 
& 0.00\% & 0.00\% & 0.00\% & 0.00\% & 0.00\%
& 0.00\% & 0.00\% & 0.00\% & 0.00\% & 0.00\% \\

\multicolumn{2}{c|}{\texttt{MI}+Multi-task SFT (\texttt{LC-Rec})}
& +6.38\% & +5.20\% & +3.87\% & +3.00\% & +3.47\%
& +5.63\% & +5.35\% & +2.98\% & +3.32\% & +3.05\% \\

\multicolumn{2}{c|}{\textbf{\texttt{GTI}+Multi-task SFT (\texttt{Ours})}} 
& {\ul \textbf{+21.63\%}} & {\ul \textbf{+13.59\%}} & {\ul \textbf{+8.16\%}} & {\ul \textbf{+6.35\%}} & {\ul \textbf{+4.25\%}}
& {\ul \textbf{+15.83\%}} & {\ul \textbf{+10.89\%}} & {\ul \textbf{+5.74\%}} & {\ul \textbf{+5.87\%}} & {\ul \textbf{+4.10\%}} \\

\midrule

\rowcolor{gray!15}
\multicolumn{2}{c|}{\color{gray}\textbf{\texttt{GTI}}: extra gain over \texttt{LC-Rec} ($\Delta$)}
& {\color{gray}+15.25\%} & {\color{gray}+8.39\%} & {\color{gray}+4.29\%} & {\color{gray}+3.35\%} & {\color{gray}+0.78\%}
& {\color{gray}+10.20\%} & {\color{gray}+5.54\%} & {\color{gray}+2.76\%} & {\color{gray}+2.55\%} & {\color{gray}+1.05\%} \\

\bottomrule
\end{tabular}
}
}
\label{table:retrieval_relative}
\vspace{-5pt}
\end{table*}

\begin{figure*}[b!]
    \centering
    \vspace{-7pt}
        \makebox[\textwidth][c]{\includegraphics[width=1.1\linewidth]{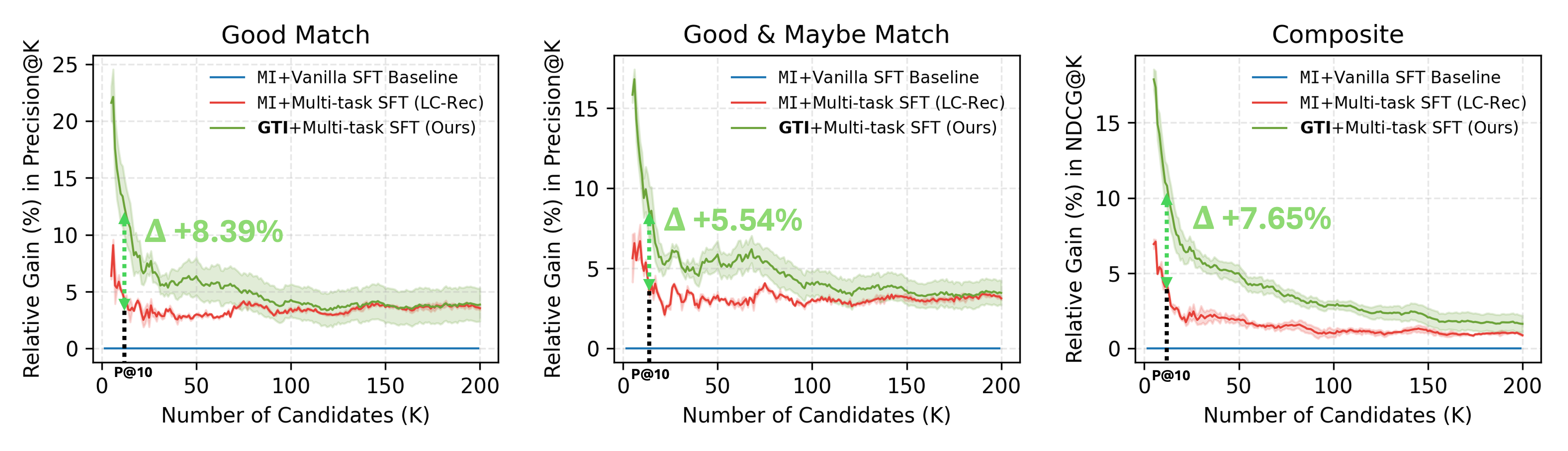}}
    \vspace{-20pt}
    \caption{\textbf{Relative gain versus candidate pool size.} \textbf{Left/Middle}: Relative Precision@K gain under Good Match and Good \& Maybe Match; \textbf{Right}: Relative NDCG@K gain (Composite). \fancyname\ consistently outperforms both baselines across all pool sizes, with the largest gains at small $K$. Shaded areas denote variability across runs.}
    \vspace{-10pt}
    \label{fig:sweep-over-k}
\end{figure*}

\paragraph{Implementation Details.}
Across both datasets, we employ Qwen3-0.6B as the backbone language model. Semantic IDs are constructed via RQ-VAE, following the formulation in \citet{TIGER}. For \fancyname, the grounding stage freezes all parameters except $E_{\text{new}}$ and trains for 8,000 steps with batch size 128; all parameters are then unfrozen for an additional 8,000 steps at the same batch size, followed by the standard SFT procedure used for the baseline. All experiments use four NVIDIA H100 GPUs.

\textit{Industrial dataset.} Candidate-level semantic representations are obtained by fine-tuning \texttt{Mistral-E5} in a two-tower architecture with recruiter engagement signals, producing 1024-dimensional embeddings. The RQ-VAE uses $L=3$ codebook levels with $K=8{,}192$ codes per level. The subsequent SFT baseline trains with a batch size of 512 for 1,600 steps.

\textit{Public dataset.} Item-level semantic representations are derived using the off-the-shelf \texttt{Qwen3-Embedding-0.6B} encoder, yielding 1024-dimensional vectors. The RQ-VAE uses a 3-layer MLP encoder--decoder with ReLU activations, $L=4$ codebook levels with $K=256$ codes per level (32-dimensional codes), and the diversity regularizer of \citet{LETTER} to encourage balanced codebook utilization. The RQ-VAE is trained for 20K epochs. The SFT baseline trains with batch size 512 for 1,600 steps.

\begin{table*}[]
\vspace{-40pt}
\centering
\caption{Relative NDCG@K (Composite) gain (\%) over SFT baseline on a real-world candidate retrieval dataset. Bold and underline denote the best result.}
\vspace{5pt}
\scriptsize
\begin{tabular}{clccccc}
\toprule
\multicolumn{2}{c|}{\multirow{2}{*}{Methodology}} & \multicolumn{5}{c}{NDCG@K (Composite)} \\
\multicolumn{2}{c|}{} & @5 & @10 & @20 & @50 & @100 \\ 
\midrule

\multicolumn{2}{c|}{\texttt{MI}+Vanilla SFT (\texttt{Baseline})}
& 0.00\% & 0.00\% & 0.00\% & 0.00\% & 0.00\% \\

\multicolumn{2}{c|}{\texttt{MI}+Multi-task SFT (\texttt{LC-Rec})}
& +6.94\% & +4.38\% & +1.94\% & +1.95\% & +1.01\% \\

\multicolumn{2}{c|}{\textbf{\texttt{GTI}+Multi-task SFT (\texttt{Ours})}}
& {\ul \textbf{+17.88\%}} & {\ul \textbf{+12.03\%}} & {\ul \textbf{+6.90\%}} & {\ul \textbf{+4.99\%}} & {\ul \textbf{+2.89\%}} \\

\midrule

\rowcolor{gray!15}
\multicolumn{2}{c|}{\color{gray}\textbf{\texttt{GTI}}: extra gain over \texttt{LC-Rec} ($\Delta$)}
& {\color{gray}+10.94\%} & {\color{gray}+7.65\%} & {\color{gray}+4.96\%} & {\color{gray}+3.04\%} & {\color{gray}+1.88\%} \\

\bottomrule
\end{tabular}
\label{table:ndcg_relative}
\vspace{-10pt}
\end{table*}

\begin{table*}[b!]
\vspace{-20pt}
\caption{Relative Recall@K and NDCG@K (\%) over SFT baseline on Vibrent Dataset.}
\vspace{5pt}
\makebox[\linewidth][c]{%
\resizebox{1.1\linewidth}{!}{
\begin{tabular}{cl|ccccc|ccccc}
\toprule
\multicolumn{2}{c|}{\multirow{2}{*}{Methodology}} 
& \multicolumn{5}{c|}{Recall@K} 
& \multicolumn{5}{c}{NDCG@K} \\
\multicolumn{2}{c|}{} 
& @5 & @10 & @20 & @50 & @100 
& @5 & @10 & @20 & @50 & @100 \\
\midrule

\multicolumn{2}{c|}{\texttt{MI}+Vanilla SFT (\texttt{Baseline})} 
& 0.00\% & 0.00\% & 0.00\% & 0.00\% & 0.00\%
& 0.00\% & 0.00\% & 0.00\% & 0.00\% & 0.00\% \\

\multicolumn{2}{c|}{\texttt{MI}+Multi-task SFT (\texttt{LC-Rec})} 
& {\ul \textbf{+7.69\%}} & +11.86\% & +13.41\% & +12.03\% & +15.73\%
& {\ul \textbf{+8.47\%}} & {\ul \textbf{+10.74\%}} & +11.30\% & +11.18\% & {\ul \textbf{+13.26\%}} \\

\multicolumn{2}{c|}{\textbf{\texttt{GTI}+Vanilla SFT (\texttt{Ours})}} 
& +1.71\% & {\ul \textbf{+22.03\%}} & {\ul \textbf{+26.02\%}} & {\ul \textbf{+21.55\%}} & {\ul \textbf{+18.54\%}}
& -5.19\% & +8.02\% & {\ul \textbf{+12.23\%}} & {\ul \textbf{+12.83\%}} & +12.46\% \\

\bottomrule
\end{tabular}
}
}
\label{tab:vibrent_results}
\vspace{-5pt}
\end{table*}

\begin{figure*}[b!]
    \centering
    \vspace{-5pt}
        \makebox[\textwidth][c]{\includegraphics[width=0.8\linewidth]{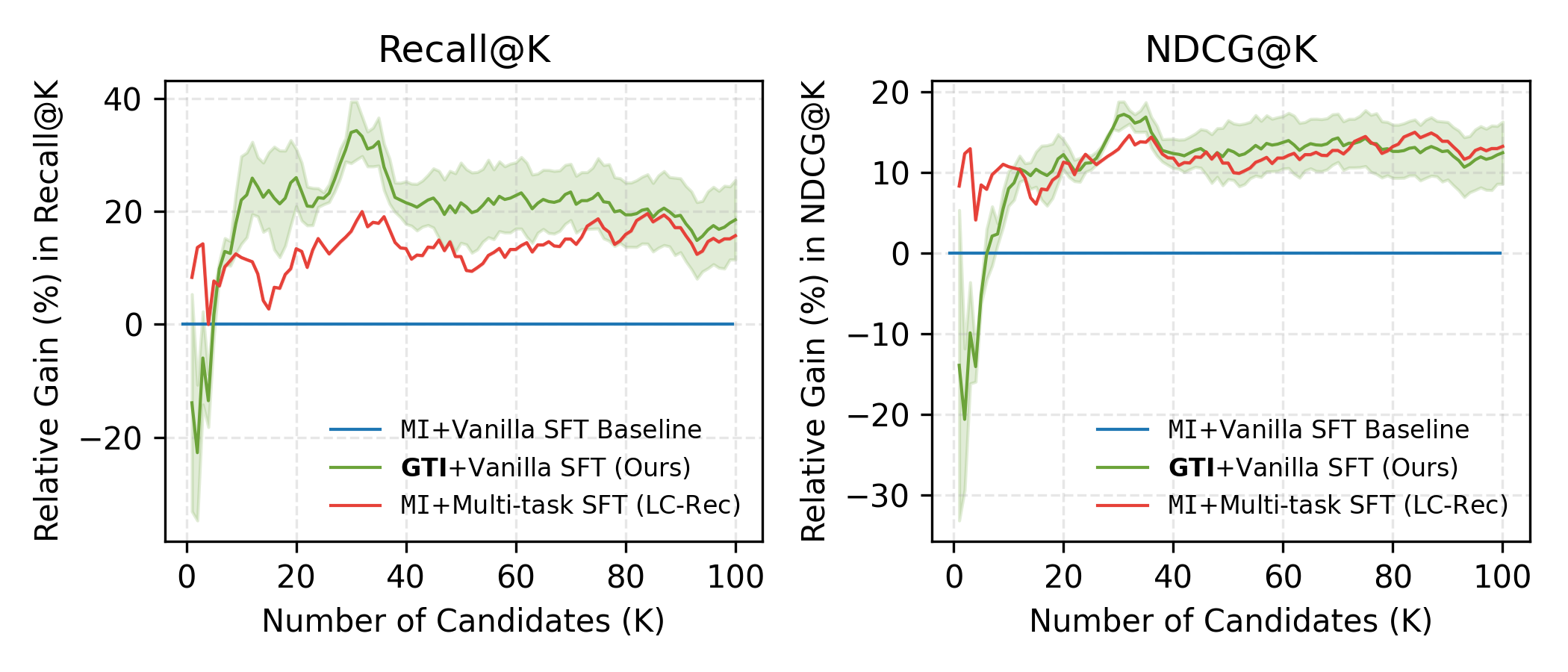}}
    \vspace{-25pt}
    \caption{\textbf{Relative gain versus candidate pool size.} \textbf{Left}: Relative Recall@K gain; \textbf{Right}: Relative NDCG@K gain. Shaded areas denote variability across runs.}
    \vspace{-15pt}
    \label{fig:vibrent-sweep-over-k}
\end{figure*}

\subsection{Overall Performance Analysis}


Tables~\ref{table:retrieval_relative} \& ~\ref{table:ndcg_relative} and Figure~\ref{fig:sweep-over-k} detail the overall performance on the industrial-scale dataset.

\textbf{The effectiveness of \fancyname\ Initialization.} Across all cutoffs, evaluation metrics, and relevance thresholds (Good Match and Good \& Maybe Match), \fancyname\ outperforms both baselines. Under the strict Good Match criterion, \fancyname\ achieves \textbf{+21.63\%} relative gain at P@5 over vanilla SFT, compared to +6.38\% for \texttt{LC-Rec}, yielding an extra gain $\Delta$ of 15.25\% attributable to the grounding stage. This pattern is consistent across evaluation settings: under Good \& Maybe Match, \fancyname\ maintains a clear advantage (\textbf{+15.83\%} vs.\ +5.63\% at P@5), and NDCG@5 exhibits the same trend (\textbf{+17.88\%} vs.\ +6.94\%). Sweeping the candidate pool size from 5 to 200 (Figure~\ref{fig:sweep-over-k}) further confirms that the improvement is robust across retrieval scales.

\textbf{Evidence for the \fancyname\ hypothesis.} The comparison between \texttt{LC-Rec} and \fancyname\ provides a controlled test of our hypothesis, as both methods introduce linguistic supervision for new tokens but differ in \emph{when} it is applied: \texttt{LC-Rec} incorporates auxiliary language modeling objectives during fine-tuning while retaining mean initialization, whereas \fancyname\ addresses the initialization directly through a grounding stage that precedes fine-tuning. The consistent performance gap ({\color{gray}extra gain $\Delta$}) between the two methods, despite sharing the same downstream SFT procedure, suggests that grounding new tokens before fine-tuning provides a more effective starting point than relying on auxiliary objectives alone, consistent with the Grounded Token Initialization hypothesis.

\paragraph{Controlled comparison on public dataset.} To disentangle the effect of grounded initialization from that of multi-task adaptation and assess the generalization of our method beyond the proprietary dataset, we compare \fancyname+Vanilla SFT against \texttt{LC-Rec} (Multi-task SFT) on the public Vibrent dataset (Table~\ref{tab:vibrent_results} and Figure~\ref{fig:vibrent-sweep-over-k}). Even without auxiliary objectives during fine-tuning, \fancyname\ achieves substantially higher Recall at $K \geq 10$ (e.g., \textbf{+26.02\%} vs.\ +13.41\% at Recall@20) and comparable NDCG, indicating that the grounding stage alone accounts for a large portion of the downstream improvement.

\subsection{Further Analysis}
\label{sec:analysis}

\begin{figure}[]
    \centering
    \vspace{-20pt}
    \includegraphics[width=0.75\linewidth]{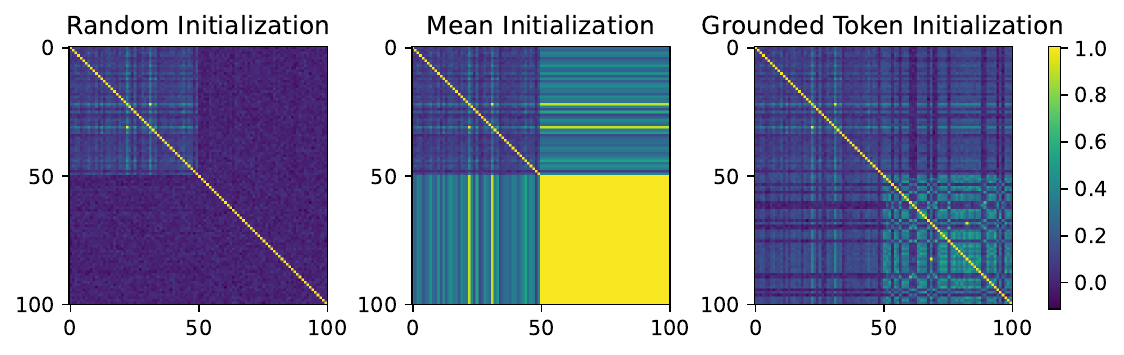}
    \vspace{-14pt}
    \caption[Pairwise cosine-similarity matrices under three initialization strategies.]{\textbf{Pairwise cosine-similarity matrices under three initialization strategies.} Each matrix shows similarities between 50 pretrained tokens (upper-left block) and 50 SID tokens (bottom-right block)\footnotemark. Random initialization (left) yields noninformative SID embeddings. Mean initialization (middle) collapses SID tokens into a near-uniform block. \texorpdfstring{\fancyname{}}{GTI} (right) produces differentiated intra-SID structure with meaningful affinities to pretrained tokens.}
    \vspace{-5pt}
    \label{fig:init_cossim}
\end{figure}

\footnotetext{For better visualization, we randomly choose 50 tokens separately from pretrained tokens or Semantic-ID tokens}

\begin{figure}[]
    \centering
    \makebox[\linewidth][c]{%
    \includegraphics[width=1.05\linewidth]{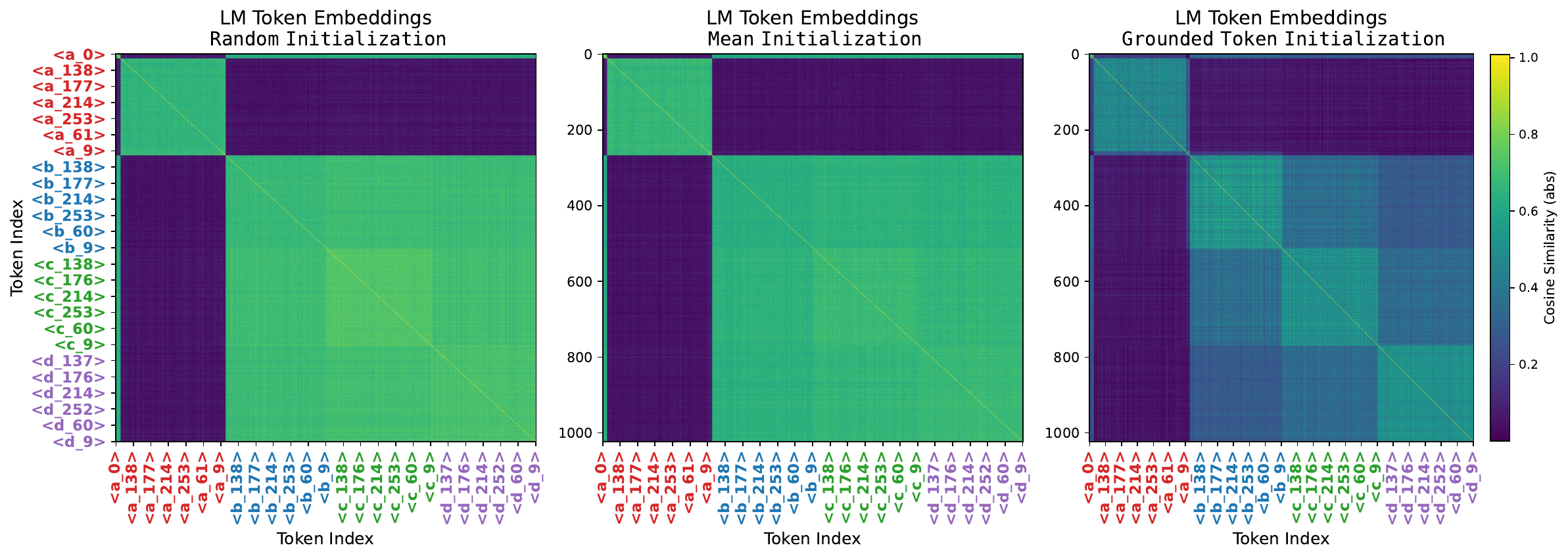}
    }
    \vspace{-15pt}
    \caption{\textbf{Pairwise SID similarity after fine-tuning (public dataset).} We visualize the pairwise cosine similarity matrix of SID embeddings at the fine-tuned checkpoint. \fancyname\ is the only initialization strategy that preserves a clear blockwise hierarchical semantics among SID tokens, suggesting improved preservation of semantic geometry. By contrast, mean and random initialization produce flat or noisy similarity patterns even after SFT stage.}
    \vspace{-10pt}
    \label{fig:pairwise_sft}
\end{figure}

The preceding results establish \emph{that} grounded initialization improves downstream performance; we now investigate \emph{why}. We use spectral and geometric diagnostics on the SID embedding subspace, both at initialization and after fine-tuning. These analyses provide direct evidence to the Grounded Token
Initialization Hypothesis (Section~\ref{subsect:misalign}).

\paragraph{Grounded initialization produces differentiated embedding geometry.} Figure~\ref{fig:init_cossim} visualizes pairwise cosine similarities among pretrained vocabulary tokens and SID tokens under three initialization strategies. Random initialization avoids uniformity but yields unstructured noise with no coherent affinity to the pretrained manifold. Mean initialization produces a uniform SID block, confirming the collapse diagnosed in Section~\ref{subsect:misalign}. In contrast, \fancyname\ produces rich, differentiated structure within the SID block together with coherent cross-block affinities to relevant lexical tokens.

\begin{figure}[h]
    \centering
    \vspace{-30pt}
    \includegraphics[width=.9\linewidth]{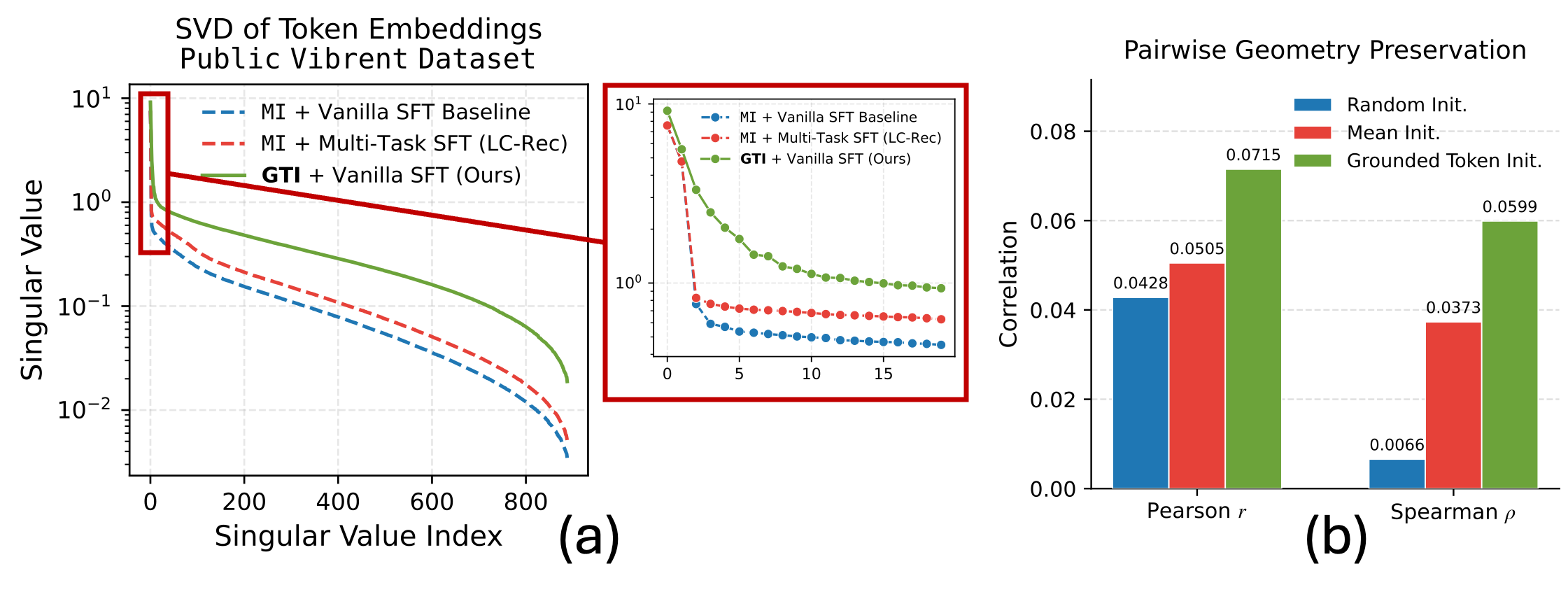}
    \vspace{-10pt}
    \caption{(a)~\textbf{Singular-Value Spectra} of SID embedding matrix after SFT: \fancyname\ initialization yields slower spectral decay and higher effective rank than mean initialization. (b)~\textbf{Representational Similarity Analysis (RSA)} of SID embeddings after SFT: We compare the pairwise geometry of the ground-truth RQ-VAE codebook vectors and the learned SID embeddings using Pearson \(r\) and Spearman \(\rho\). \fancyname\ initialization achieves the highest correlation under both metrics, indicating better preservation of the semantic structure among SID embeddings than mean or random initialization.}
    \vspace{-15pt}
    \label{fig:svd_rsa}
\end{figure}

\paragraph{Grounded structure persists through fine-tuning.} We next examine whether the structure induced by grounding persists through fine-tuning. (1)~ Pairwise cosine similarities among SID embeddings after fine-tuning on the public dataset (Figure~\ref{fig:pairwise_sft}) show that only the \fancyname-initialized model preserves the blockwise hierarchical structure encoded by the RQ-VAE; mean and random initialization produce flat or noisy similarity patterns. (2)~ The singular-value spectrum of $E_{\mathrm{SID}}\in \mathbb{R}^{|\mathcal{V}_{\text{SID}}|\times d}$ after fine-tuning on the industrial dataset (Figure~\ref{fig:svd_rsa}a) shows that mean initialization leads to rapid spectral decay and low effective rank, while grounded initialization yields slower decay and higher effective rank, indicating a non-degenerate subspace with multiple active directions along which items differ (see Appendix  for extended SVD analysis of the industrial dataset).(3)~Representational similarity analysis (RSA) between the learned SID embeddings and the ground-truth RQ-VAE codebook vectors (Figure~\ref{fig:svd_rsa}b) shows that \fancyname-initialized embeddings better preserve the original semantic structure through training. Taken together, these results suggest that the grounding stage seeds embedding structure that persists through fine-tuning, corroborating the downstream performance gains.

%% file: sections/05related_work.tex
\section{Related Work}

\paragraph{Vocabulary Extension in Language Models.}
Extending a pretrained LM's vocabulary with new tokens is a recurring challenge. Standard approaches initialize new embeddings at the vocabulary mean~\citep{hewitt2021initializing} or randomly, then rely on fine-tuning. ToolkenGPT~\citep{ToolkenGPT} and Yo'LLaVA~\citep{YoLLaVA} show that training only new token embeddings against a frozen LM can be effective for tool invocation and visual concept grounding, respectively. \fancyname\ reframes this mechanism as an \emph{initialization} strategy: by grounding new tokens before fine-tuning, the learned structure serves as a starting point that benefits arbitrary downstream tasks, rather than being tied to a specific end use.

\paragraph{Generative Recommendation.}
We adopt generative recommendation as our primary evaluation domain, as it requires injecting thousands of novel tokens into a pretrained LM, making it a demanding testbed for vocabulary extension. This paradigm frames retrieval as autoregressive decoding of Semantic IDs (SIDs) discretized via RQ-VAE~\citep{vq-vae,RQ-VAE,TIGER,LC-Rec}. We provide an extended discussion in Appendix~\ref{sec:full_relatedwork}.

%% file: sections/06conclusion.tex
\section{Conclusion}

Through spectral and geometric diagnostics, we show that mean-of-vocabulary initialization collapses new tokens into a degenerate subspace that fine-tuning does not fully recover. Motivated by this diagnosis, we propose \fancyname, a lightweight grounding stage that learns only the new token embeddings via paired linguistic supervision before standard fine-tuning. On generative recommendation benchmarks spanning industrial-scale and public datasets, \fancyname\ consistently outperforms both mean initialization and auxiliary-task adaptation, with further analyses confirming that grounded structure persists through fine-tuning. These findings support the Grounded Token Initialization Hypothesis. As the grounding mechanism makes no assumptions about the downstream task, an important direction for future work is to test its generality in broader vocabulary-extension settings beyond recommendation.

%% file: sections/07acknowledgment.tex



%% file: sections/99appendix.tex
\newpage

\section{Appendix}

\subsection{Datasets}
\label{app:dataset}
\subsubsection{Retrieval Dataset}

\paragraph{Industrial Candidate Retrieval Dataset.} The industrial-scale candidate retrieval dataset\footnote{Data access and use complied with internal privacy and security frameworks. Member profile attributes were processed in accordance with applicable member controls and visibility settings, and analyses were conducted on de-identified datasets with results reported in aggregate only.} consists of job requirement--candidate pairs collected in 2025 from a world-leading professional networking platform with global user coverage and evaluated by our internal LLM judge. According to product policy, which measures how many job requirements a candidate satisfies, each pair is assigned to one of three relevance levels: good match, good\&maybe match, and not match. We use the good match pairs for supervised fine-tuning (SFT).

The member profile dataset contains profiles of users who provide at least one of the following attributes: geographic location, job positions, education history, or skill information.

\paragraph{Vibrent Dataset.} The
\href{https://www.kaggle.com/datasets/kaborg15/vibrent-clothes-rental-dataset}{Vibrent Clothes Rental Dataset} is 
a publicly available dataset from Kaggle. To complement our industrial dataset with a publicly available benchmark, we also evaluate our method
on it.
The dataset contains anonymized user--item rental transactions from a clothing rental platform. We construct a candidate retrieval task
by treating users as queries and clothing items as candidates, where observed rental interactions are
considered positive relevance signals, and non-interacted items are treated as negatives during training.

\subsection{Prompt Templates}
\label{app:prompt-template}
\subsubsection{Prompt Template: Auxiliary Task (Item Title/Description $\leftrightarrow$ New Vocabulary Tokens)}

\begin{promptboxgray}[label={lst:semid-prompt}]{Item Title/Description $\rightarrow$ Semantic IDs\footnotemark  \textit{(Title$\rightarrow$New Vocabulary Tokens)}}
<system>
You are a helpful assistant.
<user>
Which item has the title: {{title}}?
<assistant>
{{ITEM SEMANTIC_ID}}
\end{promptboxgray}

\begin{promptboxgray}[label={lst:semid-prompt}]{Item Title/Description $\rightarrow$ Semantic IDs \textit{(Description$\rightarrow$New Vocabulary Tokens)}}
<system>
You are a helpful assistant.
<user>
Can you tell me what item is described as {{description}}?
<assistant>
{{ITEM SEMANTIC_ID}}
\end{promptboxgray}

\begin{promptboxgray}[label={lst:semid-prompt}]{Item Title/Description $\rightarrow$ Semantic IDs \textit{(Title+Description$\rightarrow$New Vocabulary Tokens)}}
<system>
You are a helpful assistant.
<user>
What item is called {{title}} and described as {{description}}?
<assistant>
{{ITEM SEMANTIC_ID}}
\end{promptboxgray}

\begin{promptboxgreen}[label={lst:semid-prompt}]{Semantic IDs $\rightarrow$ Item Title/Description \textit{(New Vocabulary Tokens$\rightarrow$Title)}}
<system>
You are a helpful assistant.
<user>
Could you please tell me what item {{ITEM SEMANTIC_ID}} is called?
<assistant>
{{title}}
\end{promptboxgreen}

\begin{promptboxgreen}[label={lst:semid-prompt}]{Semantic IDs $\rightarrow$ Item Title/Description \textit{(New Vocabulary Tokens$\rightarrow$Description)}}
<system>
You are a helpful assistant.
<user>
Briefly describe item {{ITEM SEMANTIC_ID}}.
<assistant>
{{description}}
\end{promptboxgreen}

\begin{promptboxgreen}[label={lst:semid-prompt}]{Semantic IDs $\rightarrow$ Item Title/Description \textit{(New Vocabulary Tokens$\rightarrow$Title+Description)}}
<system>
You are a helpful assistant.
<user>
What is the title and description of item {{ITEM SEMANTIC_ID}}?
<assistant>
{{title}}\n\n{{description}}
\end{promptboxgreen}

\footnotetext{Most of \textit{Item Title/Description $\leftrightarrow$ Semantic IDs} prompts and retrieval prompts are adapted from~\citep{LC-Rec}.}

\subsubsection{Prompt Template: Search Query Task}

\begin{promptboxgreen}[label={lst:semid-prompt}]{Candidate Description $\rightarrow$ Semantic Id Alignment Prompt \footnotemark} 
<user>
Instruction: Predict the candidate SID. Input:
{{candidate headline}}
{{candidate profile description}}
{{candidate job title}}
{{candidate education}}
{{candidate employment history}}
{{candidate location}}
{{candidate skills}}
{{candidate company}}
<assistant>
{{CANDIDATE SEMANTIC_ID}}
\end{promptboxgreen}

\begin{promptboxgreen}[label={lst:semid-prompt}]{Semantic Id $\rightarrow$ Candidate Description Alignment Prompt}
<user>
Instruction: Recover the candidate text. Input: {{CANDIDATE SEMANTIC_ID}}
<assistant>
{{candidate headline}}
{{candidate profile description}}
{{candidate job title}}
{{candidate education}}
{{candidate employment history}}
{{candidate location}}
{{candidate skills}}
{{candidate company}}
\end{promptboxgreen}

\begin{promptboxgreen}[label={lst:semid-prompt}]{Search Query $\rightarrow$ Candidate Semantic Id Alignment Prompt}
<user>
Instruction: Predict the relevant candidate SID that matches job requirements as much as possible. If there are no candidates that meet the criteria, return the candidate SID that prioritizes the matching required requirements then matching the preferred requirements. Input: 
{{required requirements}}
{{preferred requirements}}
<assistant>
{{ITEM SEMANTIC_ID}}
\end{promptboxgreen}
\footnotetext{For brevity, we illustrate only three representative prompting templates.}

\subsubsection{Prompt Template: Retrieval Task}

\begin{promptboxgray}[label={lst:semid-prompt}]{Retrieval Prompt (Template 1)}
<system>
You are a helpful assistant.
<user>
The user has interacted with items {{inters}} in chronological order. Can you predict the next possible item that the user may expect?
<assistant>
{{ITEM SEMANTIC_ID}}
\end{promptboxgray}

\begin{promptboxgray}[label={lst:semid-prompt}]{Retrieval Prompt (Template 2)}
<system>
You are a helpful assistant.
<user>
Based on the items that the user has interacted with: {{inters}}, can you determine what item would be recommended to the user next?
<assistant>
{{ITEM SEMANTIC_ID}}
\end{promptboxgray}

\begin{promptboxgray}[label={lst:semid-prompt}]{Retrieval Prompt (Template 3)}
<system>
You are a helpful assistant.
<user>
Here is the item interaction history of the user: {{inters}}, what to recommend to the user next?
<assistant>
{{ITEM SEMANTIC_ID}}
\end{promptboxgray}

\subsection{Implementation Details}
\label{app:hyperparams}

We utilize the pre-trained \texttt{Qwen3-Embedding-0.6B} encoder to extract semantic representations for items. The encoder processes item metadata including titles and descriptions to generate 1024-dimensional dense vectors that capture semantic similarities between items.
We process text features of products by concatenating them as: [TITLE] [DESCRIPTION]. We set the maximum input sequence length as 2048. The final outputs are dense semantic embeddings: $z_i\in\mathbb{R}^{1024}$ for item $i$.

Our Residual Quantized Variational Autoencoder (RQ-VAE) follows the TIGER~\citep{TIGER} framework with carefully designed architectural specifications to ensure effective quantization of semantic representations. The encoder architecture consists of a 3-layer Multi-Layer Perceptron (MLP) with hidden dimensions of [1024, 512, 256], utilizing ReLU activation functions and applying a dropout rate of 0.1 between layers. The residual quantization mechanism employs four codebook layers, each containing 256 entries with 32-dimensional codes. This hierarchical quantization approach enables fine-grained representation of semantic information while maintaining discrete tokenization properties essential for language model integration. We trained the model for 20,000 epochs to achieve a high codebook utilization rate and minimize collision rates. To further prevent collisions where multiple items map to identical sequences of semantic IDs, we employed the Sinkhorn-Knopp trick used by LC-Rec~\citep{LC-Rec}, which ensures uniform distribution of item semantics across codebook embeddings in the final layer.

The base language model employs \texttt{Qwen3-0.6B} with hidden dimension of 1024. The model architecture comprises 28 transformer layers supporting a maximum context length of 32,768 tokens. This configuration provides sufficient capacity for processing sequential recommendation tasks while maintaining computational efficiency. Parameter-efficient fine-tuning is implemented through Quantized Low-Rank Adaptation (QLoRA) with a rank of 8 and alpha value of 32. The LoRA adaptation applies a dropout rate of 0.05 and targets key projection matrices including q\_proj, k\_proj, v\_proj, o\_proj, gate\_proj, up\_proj, and down\_proj. We also set LoRA modules to be saved as embed\_tokens and lm\_head, so that only the embedding layer and the language modeling head are preserved during training while other modules can remain frozen. This configuration enables efficient adaptation while preserving pre-trained knowledge.

We implement the token-embedding grounding stage of \fancyname\ by extending the Hugging Face TRL~\citep{huggingface_trl_quickstart} \texttt{SFTTrainer} to update only the Semantic-ID embedding matrix while freezing the LM backbone; the trainer consumes paired (title/description, SemID) examples and optimizes the embeddings as outlined in the pseudo code below. Unless otherwise stated, we train for 10 epochs with a learning rate of 1e-3 and a batch size 16.

\begin{algorithm*}[]
\caption{\fancyname\ Grounding Stage}
\label{alg:gti-grounding}
\DontPrintSemicolon
\KwIn{Pretrained model $\mathcal{M}$ with embedding matrix $E \in \mathbb{R}^{V \times d}$; new token indices $\mathcal{T} \subseteq \{0,\ldots,V-1\}$; paired corpus $\mathcal{D} = \{(\text{text}_j, \text{token}_j)\}$}
\KwOut{Model $\mathcal{M}$ with grounded embeddings for tokens in $\mathcal{T}$}
\BlankLine

\tcp{Setup: freeze all parameters except new token embeddings}
\BlankLine
Freeze all parameters of $\mathcal{M}$ \;
Construct binary mask $\mathbf{m} \in \{0,1\}^V$ where $m_i = 1$ iff $i \in \mathcal{T}$ \;
$\mathbf{M} \leftarrow \mathbf{m} \otimes \mathbf{1}_d$ \tcp*{Broadcast to $\mathbb{R}^{V \times d}$}
\BlankLine
\BlankLine

\tcp{Training: update only new token embeddings via masked gradients}
\BlankLine
\For{each batch $\mathcal{B} \subset \mathcal{D}$}{
    $\mathcal{L} \leftarrow \textsc{LM\_Loss}(\mathcal{M}, \mathcal{B})$ \tcp*{Forward pass}
    $\nabla E \leftarrow \nabla_E \mathcal{L}$ \tcp*{Compute gradients}
    $E \leftarrow E - \eta \cdot (\nabla E \odot \mathbf{M})$ \tcp*{Update only new token embeddings}
}
\end{algorithm*}

\subsection{Analysis Details}
\label{app:analysis}

\begin{table*}[b!]
\vspace{-20pt}
\caption{\textbf{Additional retrieval results on the Vibrent dataset.} 
We report Recall@K and NDCG@K for \texttt{Baseline} (\texttt{MI}+Vanilla SFT), 
\texttt{LC-Rec} (\texttt{MI}+Multi-task SFT), and our method 
\texttt{GTI}+Vanilla SFT. \texttt{GTI}+Vanilla SFT achieves the best performance 
on most Recall@K metrics and remains competitive on NDCG@K, further supporting 
the effectiveness of grounded token initialization for generative retrieval.}
\vspace{5pt}
\makebox[\linewidth][c]{%
\resizebox{1.1\linewidth}{!}{
\begin{tabular}{cl|ccccc|ccccc}
\toprule
\multicolumn{2}{c|}{\multirow{2}{*}{Methodology}} 
& \multicolumn{5}{c|}{Recall@K} 
& \multicolumn{5}{c}{NDCG@K} \\
\multicolumn{2}{c|}{} 
& @5 & @10 & @20 & @50 & @100 
& @5 & @10 & @20 & @50 & @100 \\
\midrule

\multicolumn{2}{c|}{\texttt{MI}+Vanilla SFT (\texttt{Baseline})} 
& 0.0226 & 0.0342 & 0.0475 & 0.0771 & 0.1031
& 0.0150 & 0.0188 & 0.0222 & 0.0280 & 0.0322 \\

\multicolumn{2}{c|}{\texttt{MI}+Multi-task SFT (\texttt{LC-Rec})} 
& \textbf{0.0243} & 0.0382 & 0.0539 & 0.0863 & 0.1194
& \textbf{0.0163} & \textbf{0.0208} & 0.0247 & 0.0311 & \textbf{0.0365} \\

\multicolumn{2}{c|}{\textbf{\texttt{GTI}+Vanilla SFT (\texttt{Ours})}} 
& 0.0230 & \textbf{0.0417} & \textbf{0.0599} & \textbf{0.0937} & \textbf{0.1222}
& 0.0143 & 0.0203 & \textbf{0.0249} & \textbf{0.0316} & 0.0362 \\

\bottomrule
\end{tabular}
}
}
\label{tab:vibrent_results}
\vspace{-5pt}
\end{table*}

\paragraph{Representation Similarity Analysis (RSA).} To quantitatively measure whether the learned representations preserves the semantic structure of SID new vocabulary tokens, we perform representational similarity analysis. Given the well-trained RQ-VAE codebooks, which encode the compressed representation of SID new vocabulary tokens, we define the oracle semantic embeddings as $X=\{x_1,...,x_n\}, x_i\in\mathbf{R}^{32}$. And let the corresponding learned token embeddings from language model as $\hat{X}=\{\hat{x}_1,...,\hat{x}_n\},x_i\in\mathbf{R}^{d}$, where $d$ depends on the language model dimensionality. We construct pairwise token similarity matrices $S_X, S_{\hat{X}}\in\mathbf{R}^{n\times n}$, where: 
$$
(S_X)_{i,j}=\cos(x_i,x_j),\qquad (S_{\hat{X}})_{i,j}=\cos(\hat{x}_i,\hat{x}_j).
$$
We then vectorize the upper-triangular entries of $S_X$ and $S_{\hat{X}}$ and compute their correlation (We implement both Spearman correlation and Pearson correlation to capture complementary aspects of representational alignment). This yields an RSA score that quantifies the extent to which the learned representation space preserves the pairwise semantic relations of the oracle space. Since RSA compares representational geometry rather than coordinates directly, it is well suited to our setting where the oracle and learned embeddings live in different ambient dimensions ($32$ vs. $d$).

\paragraph{Extended SVD analysis of the industrial dataset.}

The slower spectral decay and higher effective rank observed with \fancyname\ initialization suggest that this method preserves a more expressive and diverse feature space throughout the SFT process, preventing the dimensional collapse often associated with mean initialization.

\begin{figure*}[h!]
    \centering
    \includegraphics[width=.6\linewidth]{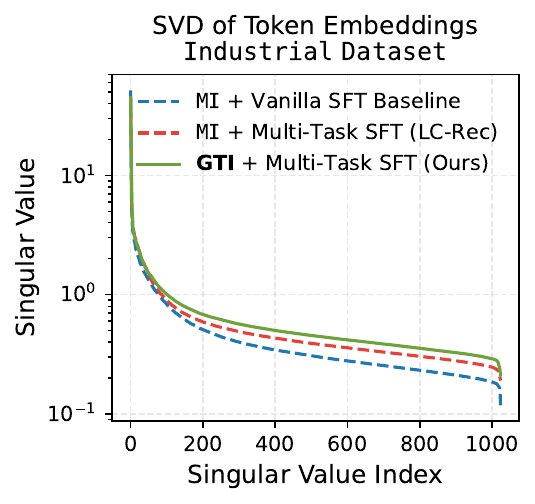}
    \caption{Singular-Value Spectra of SID embedding matrix after SFT for Industrial dataset.}
    \label{fig:svd_industrial}
\end{figure*}

\subsection{Full Related Work}
\label{sec:full_relatedwork}

\paragraph{RQ-VAE and Semantic IDs.} Vector-quantized autoencoders~\citep{vq-vae, haichao} learn discrete item representations by mapping continuous embeddings to codebook entries. Residual Quantized VAEs (RQ-VAE)~\citep{RQ-VAE} extend this with a hierarchy of residual codebooks, producing multi-level Semantic IDs (SIDs) that capture progressively finer semantic distinctions. Unlike conventional item IDs, SIDs carry compositional structure amenable to autoregressive generation, making them a standard component in generative recommendation~\citep{TIGER,LC-Rec,mtgr}. Crucially, each codebook entry becomes a new token in the LM vocabulary, and how these tokens are initialized is precisely the bottleneck our work addresses.

\paragraph{Generative Recommendation.}
Generative retrieval reframes recommendation as autoregressive decoding of item identifiers rather than nearest-neighbor search in embedding space~\citep{Aniket,chen2025pal}. TIGER~\citep{TIGER} introduced RQ-VAE-learned SIDs as generation targets, and LC-Rec~\citep{LC-Rec} added auxiliary linguistic objectives during fine-tuning to improve SID representations. Several systems have demonstrated industrial-scale deployment: MTGR~\citep{mtgr} integrates generative retrieval with DLRM cross-feature signals; OneSearch~\citep{oneSearch} combines keyword-enhanced quantization with preference-aware rewards; and OneRec~\citep{OneRec,OneRec_TechReport} unifies retrieval and ranking via session-wise generation. Complementary directions include LLM-driven knowledge-graph recommenders~\citep{cai2025boosting} and MLLM-based world-knowledge integration~\citep{zhang2025linkedout}. All of these systems must inject novel tokens into a pretrained LM; our work addresses a step that is upstream of and complementary to their contributions, namely how those tokens should be initialized.

\paragraph{Connection to Dimensional Collapse.}
The initialization collapse we diagnose is related to \emph{dimensional collapse} in contrastive and self-supervised learning~\citep{jing2021understanding, jiang2024hard}, where learned representations are restricted to a low-dimensional subspace, eliminating fine-grained distinctions (Figure~\ref{fig:misalignment_diagram}). Mean-of-vocabulary initialization induces a similar effect: all new tokens start at the same point, forming a rank-deficient configuration. \citet{jiang2024hard} show that appropriate initialization can mitigate dimensional collapse in contrastive learning, which parallels our finding that grounding new tokens before fine-tuning preserves a higher-rank, more differentiated embedding subspace.

%% file: colm2026_conference.bib
@misc{LETTER,
      title={Learnable Item Tokenization for Generative Recommendation}, 
      author={Wenjie Wang and Honghui Bao and Xinyu Lin and Jizhi Zhang and Yongqi Li and Fuli Feng and See-Kiong Ng and Tat-Seng Chua},
      year={2024},
      eprint={2405.07314},
      archivePrefix={arXiv},
      primaryClass={cs.IR},
      url={https://arxiv.org/abs/2405.07314}, 
}

@article{zhang2025linkedout,
  title={LinkedOut: Linking World Knowledge Representation Out of Video LLM for Next-Generation Video Recommendation},
  author={Zhang, Haichao and Lu, Yao and Wang, Lichen and Li, Yunzhe and Chen, Daiwei and Xu, Yunpeng and Fu, Yun},
  journal={arXiv preprint arXiv:2512.16891},
  year={2025}
}

@misc{LC-Rec,
      title={Adapting Large Language Models by Integrating Collaborative Semantics for Recommendation}, 
      author={Bowen Zheng and Yupeng Hou and Hongyu Lu and Yu Chen and Wayne Xin Zhao and Ming Chen and Ji-Rong Wen},
      year={2024},
      eprint={2311.09049},
      archivePrefix={arXiv},
      primaryClass={cs.IR},
      url={https://arxiv.org/abs/2311.09049}, 
}

@misc{TIGER,
      title={Recommender Systems with Generative Retrieval}, 
      author={Shashank Rajput and Nikhil Mehta and Anima Singh and Raghunandan H. Keshavan and Trung Vu and Lukasz Heldt and Lichan Hong and Yi Tay and Vinh Q. Tran and Jonah Samost and Maciej Kula and Ed H. Chi and Maheswaran Sathiamoorthy},
      year={2023},
      eprint={2305.05065},
      archivePrefix={arXiv},
      primaryClass={cs.IR},
      url={https://arxiv.org/abs/2305.05065}, 
}

@misc{RQ-VAE,
      title={Autoregressive Image Generation using Residual Quantization}, 
      author={Doyup Lee and Chiheon Kim and Saehoon Kim and Minsu Cho and Wook-Shin Han},
      year={2022},
      eprint={2203.01941},
      archivePrefix={arXiv},
      primaryClass={cs.CV},
      url={https://arxiv.org/abs/2203.01941}, 
}

@misc{LightGCN,
      title={LightGCN: Simplifying and Powering Graph Convolution Network for Recommendation}, 
      author={Xiangnan He and Kuan Deng and Xiang Wang and Yan Li and Yongdong Zhang and Meng Wang},
      year={2020},
      eprint={2002.02126},
      archivePrefix={arXiv},
      primaryClass={cs.IR},
      url={https://arxiv.org/abs/2002.02126}, 
}

@misc{ToolkenGPT,
      title={ToolkenGPT: Augmenting Frozen Language Models with Massive Tools via Tool Embeddings}, 
      author={Shibo Hao and Tianyang Liu and Zhen Wang and Zhiting Hu},
      year={2024},
      eprint={2305.11554},
      archivePrefix={arXiv},
      primaryClass={cs.CL},
      url={https://arxiv.org/abs/2305.11554}, 
}

@article{mtgr,
  title={MTGR: Industrial-Scale Generative Recommendation Framework in Meituan},
  author={Han, Ruidong and Yin, Bin and Chen, Shangyu and Jiang, He and Jiang, Fei and Li, Xiang and Ma, Chi and Huang, Mincong and Li, Xiaoguang and Jing, Chunzhen and others},
  journal={arXiv preprint arXiv:2505.18654},
  year={2025}
}

@misc{GRreview,
      title={A Review of Modern Recommender Systems Using Generative Models (Gen-RecSys)}, 
      author={Yashar Deldjoo and Zhankui He and Julian McAuley and Anton Korikov and Scott Sanner and Arnau Ramisa and René Vidal and Maheswaran Sathiamoorthy and Atoosa Kasirzadeh and Silvia Milano},
      year={2024},
      eprint={2404.00579},
      archivePrefix={arXiv},
      primaryClass={cs.IR},
      url={https://arxiv.org/abs/2404.00579}, 
}

@article{MF,
  title   = {Matrix Factorization Techniques for Recommender Systems},
  author  = {Koren, Yehuda and Bell, Robert and Volinsky, Chris},
  journal = {Computer},
  volume  = {42},
  number  = {8},
  pages   = {30--37},
  year    = {2009},
  publisher = {IEEE Computer Society},
  doi     = {10.1109/MC.2009.263}
}

@inproceedings{NCF,
  title     = {Neural Collaborative Filtering},
  author    = {He, Xiangnan and Liao, Lizi and Zhang, Hanwang and Nie, Liqiang and Hu, Xia and Chua, Tat-Seng},
  booktitle = {Proceedings of the 26th International Conference on World Wide Web (WWW '17)},
  pages     = {173--182},
  year      = {2017},
  doi       = {10.1145/3038912.3052569}
}

@inproceedings{NGCF,
  title     = {Neural Graph Collaborative Filtering},
  author    = {Wang, Xiang and He, Xiangnan and Wang, Meng and Feng, Fuli and Chua, Tat-Seng},
  booktitle = {Proceedings of the 42nd International ACM SIGIR Conference on Research and Development in Information Retrieval (SIGIR '19)},
  pages     = {165--174},
  year      = {2019},
  doi       = {10.1145/3331184.3331267}
}

@InProceedings{pmlr-v235-zhai24a,
  title = 	 {Actions Speak Louder than Words: Trillion-Parameter Sequential Transducers for Generative Recommendations},
  author =       {Zhai, Jiaqi and Liao, Lucy and Liu, Xing and Wang, Yueming and Li, Rui and Cao, Xuan and Gao, Leon and Gong, Zhaojie and Gu, Fangda and He, Jiayuan and Lu, Yinghai and Shi, Yu},
  booktitle = 	 {Proceedings of the 41st International Conference on Machine Learning},
  pages = 	 {58484--58509},
  year = 	 {2024},
  editor = 	 {Salakhutdinov, Ruslan and Kolter, Zico and Heller, Katherine and Weller, Adrian and Oliver, Nuria and Scarlett, Jonathan and Berkenkamp, Felix},
  volume = 	 {235},
  series = 	 {Proceedings of Machine Learning Research},
  month = 	 {21--27 Jul},
  publisher =    {PMLR},
  url = 	 {https://proceedings.mlr.press/v235/zhai24a.html}
}

@misc{oneSearch,
      title={OneSearch: A Preliminary Exploration of the Unified End-to-End Generative Framework for E-commerce Search}, 
      author={Ben Chen and Xian Guo and Siyuan Wang and Zihan Liang and Yue Lv and Yufei Ma and Xinlong Xiao and Bowen Xue and Xuxin Zhang and Ying Yang and Huangyu Dai and Xing Xu and Tong Zhao and Mingcan Peng and Xiaoyang Zheng and Chao Wang and Qihang Zhao and Zhixin Zhai and Yang Zhao and Bochao Liu and Jingshan Lv and Jing Chen and Xiao Liang and Yuqing Ding and Chenyi Lei and Wenwu Ou and Han Li and Kun Gai},
      year={2025},
      eprint={2509.03236},
      archivePrefix={arXiv},
      primaryClass={cs.IR},
      url={https://arxiv.org/abs/2509.03236}, 
}

@INPROCEEDINGS{SelfAttnSeqRec,
  author={Kang, Wang-Cheng and McAuley, Julian},
  booktitle={2018 IEEE International Conference on Data Mining (ICDM)}, 
  title={Self-Attentive Sequential Recommendation}, 
  year={2018},
  volume={},
  number={},
  pages={197-206},
  doi={10.1109/ICDM.2018.00035}}

@misc{P5,
      title={Recommendation as Language Processing (RLP): A Unified Pretrain, Personalized Prompt \& Predict Paradigm (P5)}, 
      author={Shijie Geng and Shuchang Liu and Zuohui Fu and Yingqiang Ge and Yongfeng Zhang},
      year={2023},
      eprint={2203.13366},
      archivePrefix={arXiv},
      primaryClass={cs.IR},
      url={https://arxiv.org/abs/2203.13366}, 
}

@misc{hewitt2021initializing,
  author       = {John Hewitt},
  title        = {Initializing New Word Embeddings for Pretrained Language Models},
  year         = {2021},
  howpublished = {\url{https://nlp.stanford.edu/~johnhew/vocab-expansion.html}},
}

@inproceedings{Aniket,
 author = {Rege, Aniket and Kusupati, Aditya and S, Sharan Ranjit and Fan, Alan and Cao, Qingqing and Kakade, Sham and Jain, Prateek and Farhadi, Ali},
 booktitle = {Advances in Neural Information Processing Systems},
 editor = {A. Oh and T. Naumann and A. Globerson and K. Saenko and M. Hardt and S. Levine},
 pages = {76311--76335},
 publisher = {Curran Associates, Inc.},
 title = {AdANNS: A Framework for Adaptive Semantic Search},
 url = {https://proceedings.neurips.cc/paper_files/paper/2023/file/f062da1973ac9ac61fc6d44dd7fa309f-Paper-Conference.pdf},
 volume = {36},
 year = {2023}
}

@inproceedings{
            chen2025pal,
            title={PAL: Sample-Efficient Personalized Reward Modeling for Pluralistic Alignment},
            author={Daiwei Chen and Yi Chen and Aniket Rege and Zhi Wang and Ramya Korlakai Vinayak},
            booktitle={The Thirteenth International Conference on Learning Representations},
            year={2025},
            url={https://openreview.net/forum?id=1kFDrYCuSu},
            }

@article{OneRec,
  title        = {OneRec: Unifying Retrieve and Rank with Generative Recommender and Iterative Preference Alignment},
  author       = {Deng, Jiaxin and Wang, Shiyao and Cai, Kuo and Ren, Lejian and Hu, Qigen and Ding, Weifeng and Luo, Qiang and Zhou, Guorui},
  journal      = {arXiv preprint arXiv:2502.18965},
  year         = {2025},
  url          = {https://arxiv.org/abs/2502.18965},
}

@article{OneRec_TechReport,
  title        = {OneRec Technical Report},
  author       = {Zhou, Guorui and Deng, Jiaxin and Zhang, Jinghao and Cai, Kuo and Ren, Lejian and Luo, Qiang and Wang, Qigen and Hu, Qianqian and Huang, Rui and Wang, Shiyao and others},
  journal      = {arXiv preprint arXiv:2506.13695},
  year         = {2025},
  url          = {https://arxiv.org/abs/2506.13695},
}

@misc{vq-vae,
      title={Neural Discrete Representation Learning}, 
      author={Aaron van den Oord and Oriol Vinyals and Koray Kavukcuoglu},
      year={2018},
      eprint={1711.00937},
      archivePrefix={arXiv},
      primaryClass={cs.LG},
      url={https://arxiv.org/abs/1711.00937}, 
}

@inproceedings{
haichao,
title={{VQT}oken: Neural Discrete Token Representation Learning for Extreme Token Reduction in Video Large Language Models},
author={Haichao Zhang and Yun Fu},
booktitle={The Thirty-ninth Annual Conference on Neural Information Processing Systems},
year={2025},
url={https://openreview.net/forum?id=X8oEu4Gs3W}
}

@article{cai2025boosting, title={Boosting Knowledge Graph-based Recommendations through Confidence-Aware Augmentation with Large Language Models}, author={Cai, Rui and Wang, Chao and Cai, Qianyi and Shen, Dazhong and Xiong, Hui}, journal={arXiv preprint arXiv:2502.03715}, year={2025} }

@misc{huggingface_trl_quickstart,
  author       = {HuggingFace},
  title        = {TRL Documentation: Quickstart},
  year         = {2025},
  howpublished = {\url{https://huggingface.co/docs/trl/en/quickstart}},
  note         = {Accessed: 2025-09-23}
}

@article{jing2021understanding,
  title={Understanding dimensional collapse in contrastive self-supervised learning},
  author={Jing, Li and Vincent, Pascal and LeCun, Yann and Tian, Yuandong},
  journal={arXiv preprint arXiv:2110.09348},
  year={2021}
}

@article{jiang2024hard,
  title={Hard-Negative Sampling for Contrastive Learning: Optimal Representation Geometry and Neural-vs Dimensional-Collapse},
  author={Jiang, Ruijie and Nguyen, Thuan and Aeron, Shuchin and Ishwar, Prakash},
  journal={Transactions on Machine Learning Research},
  year={2024}
}

@inproceedings{tay2022transformer,
  title={Transformer Memory as a Differentiable Search Index},
  author={Tay, Yi and Tran, Vinh and Dehghani, Mostafa and Ni, Jianmo and Bahri, Dara and Mehta, Harsh and Qin, Zhen and Hui, Kai and Zhao, Zhe and Gupta, Jai and others},
  booktitle={Advances in Neural Information Processing Systems},
  year={2022}
}

@inproceedings{gao2019representation,
  title={Representation Degeneration Problem in Training Natural Language Generation Models},
  author={Gao, Jun and He, Di and Tan, Xu and Qin, Tao and Wang, Liwei and Liu, Tie-Yan},
  booktitle={International Conference on Learning Representations},
  year={2019}
}

@inproceedings{levy2014neural,
  title={Neural Word Embedding as Implicit Matrix Factorization},
  author={Levy, Omer and Goldberg, Yoav},
  booktitle={Advances in Neural Information Processing Systems},
  volume={27},
  year={2014}
}

@inproceedings{wolf2020transformers,
  title={Transformers: State-of-the-Art Natural Language Processing},
  author={Wolf, Thomas and Debut, Lysandre and Sanh, Victor and Chaumond, Julien and Delangue, Clement and Moi, Anthony and Cistac, Pierric and Rault, Tim and Louf, Remi and Funtowicz, Morgan and Davison, Joe and Shleifer, Sam and von Platen, Patrick and Ma, Clara and Jernite, Yacine and Plu, Julien and Xu, Canwen and Le Scao, Teven and Gugger, Sylvain and Drame, Mariama and Lhoest, Quentin and Rush, Alexander},
  booktitle={Proceedings of the 2020 Conference on Empirical Methods in Natural Language Processing: System Demonstrations},
  pages={38--45},
  year={2020}
}

@misc{YoLLaVA,
      title={Yo'LLaVA: Your Personalized Language and Vision Assistant}, 
      author={Thao Nguyen and Haotian Liu and Yuheng Li and Mu Cai and Utkarsh Ojha and Yong Jae Lee},
      year={2024},
      eprint={2406.09400},
      archivePrefix={arXiv},
      primaryClass={cs.CV},
      url={https://arxiv.org/abs/2406.09400}, 
}

@misc{vibrent_clothes_rental_dataset,
  author       = {Karl Audun Kagnes Borgersen},
  title        = {Vibrent Clothes Rental Dataset},
  year         = {2024},
  howpublished = {\url{https://www.kaggle.com/datasets/kaborg15/vibrent-clothes-rental-dataset}},
  note         = {Kaggle dataset, accessed March 22, 2026}
}

@misc{ding2026doesgenerativerecommendationgeneralize,
      title={How Well Does Generative Recommendation Generalize?}, 
      author={Yijie Ding and Zitian Guo and Jiacheng Li and Letian Peng and Shuai Shao and Wei Shao and Xiaoqiang Luo and Luke Simon and Jingbo Shang and Julian McAuley and Yupeng Hou},
      year={2026},
      eprint={2603.19809},
      archivePrefix={arXiv},
      primaryClass={cs.IR},
      url={https://arxiv.org/abs/2603.19809}, 
}
